\documentclass[10pt,twocolumn,letterpaper]{article}

\usepackage[pagenumbers]{cvpr} 

\usepackage{amsmath, amssymb, caption, subcaption, multirow, overpic, textpos,bm,graphicx, booktabs}

\usepackage[table]{xcolor}
\usepackage[british, english, american]{babel}
\usepackage[pagebackref=false, breaklinks=true, letterpaper=true, colorlinks,
            citecolor=citecolor, linkcolor=linkcolor, bookmarks=false]{hyperref}
\definecolor{citecolor}{HTML}{0071BC}
\definecolor{linkcolor}{HTML}{ED1C24}

\def\cvprPaperID{**}
\def\confName{****}
\def\confYear{****}

\newlength\savewidth\newcommand\shline{\noalign{\global\savewidth\arrayrulewidth
  \global\arrayrulewidth 1pt}\hline\noalign{\global\arrayrulewidth\savewidth}}
\newcommand{\tablestyle}[2]{\setlength{\tabcolsep}{#1}\renewcommand{\arraystretch}{#2}\centering\footnotesize}
\renewcommand{\paragraph}[1]{\vspace{1.25mm}\noindent\textbf{#1}}

\newcolumntype{x}[1]{>{\centering\arraybackslash}p{#1pt}}
\newcolumntype{y}[1]{>{\raggedright\arraybackslash}p{#1pt}}
\newcolumntype{z}[1]{>{\raggedleft\arraybackslash}p{#1pt}}

\newcommand{\app}{\raise.17ex\hbox{$\scriptstyle\sim$}}

\definecolor{deemph}{gray}{0.6}
\newcommand{\gc}[1]{\textcolor{deemph}{#1}}
\definecolor{baselinecolor}{gray}{.9}
\newcommand{\baseline}[1]{\cellcolor{baselinecolor}{#1}}

\begin{document}
\title{\Large CLIP Itself is a Strong Fine-tuner: Achieving 85.7\% and 88.0\% Top-1 Accuracy with ViT-B and ViT-L on ImageNet}

\author{Xiaoyi Dong$^{1}$ \thanks{Work done during an internship at Microsoft Research Asia}, Jianmin Bao$^{2}$,  Ting Zhang$^{2}$,  Dongdong Chen$^{3}$ , Shuyang Gu$^{2}$, \\ Weiming Zhang$^{1}$, Lu Yuan$^{3}$,  Dong Chen$^{2}$, Fang Wen$^{2}$,   Nenghai Yu$^{1}$\\
	$^{1}$University of Science and Technology of China  \\
	$^{2}$Microsoft Research Asia
	$^{3}$Microsoft Cloud $+$ AI \\
	{\tt\small\{dlight@mail, zhangwm@, ynh@\}.ustc.edu.cn } 
	{\tt\small cddlyf@gmail.com }\\
	{\tt\small\{jianbao,tinzhan,shuyanggu,luyuan,doch,fangwen\}@microsoft.com }
}
\maketitle

\begin{abstract}
	Recent studies have shown that CLIP has achieved remarkable success in performing zero-shot inference while its fine-tuning performance is not satisfactory. In this paper, we identify that fine-tuning performance is significantly impacted by hyper-parameter choices. We examine various key hyper-parameters and empirically evaluate their impact in fine-tuning CLIP for classification tasks through a comprehensive study. We find that the fine-tuning performance of CLIP is substantially underestimated. Equipped with hyper-parameter refinement, we demonstrate CLIP itself is better or at least competitive in fine-tuning compared with large-scale supervised pre-training approaches or latest works that use CLIP as prediction targets in Masked Image Modeling. Specifically, CLIP ViT-Base/16  and CLIP ViT-Large/14 can achieve $\bm{85.7\%},\bm{88.0\%}$ finetuning Top-1 accuracy on the ImageNet-1K dataset . These observations challenge the conventional conclusion that CLIP is not suitable for fine-tuning, and motivate us to rethink recently proposed improvements based on CLIP. We will release our code publicly at \url{https://github.com/LightDXY/FT-CLIP}.
	
\end{abstract}

\section{Introduction}

\begin{figure}[t]\centering
	\vspace{-1em}
	\includegraphics[width=.95\linewidth]{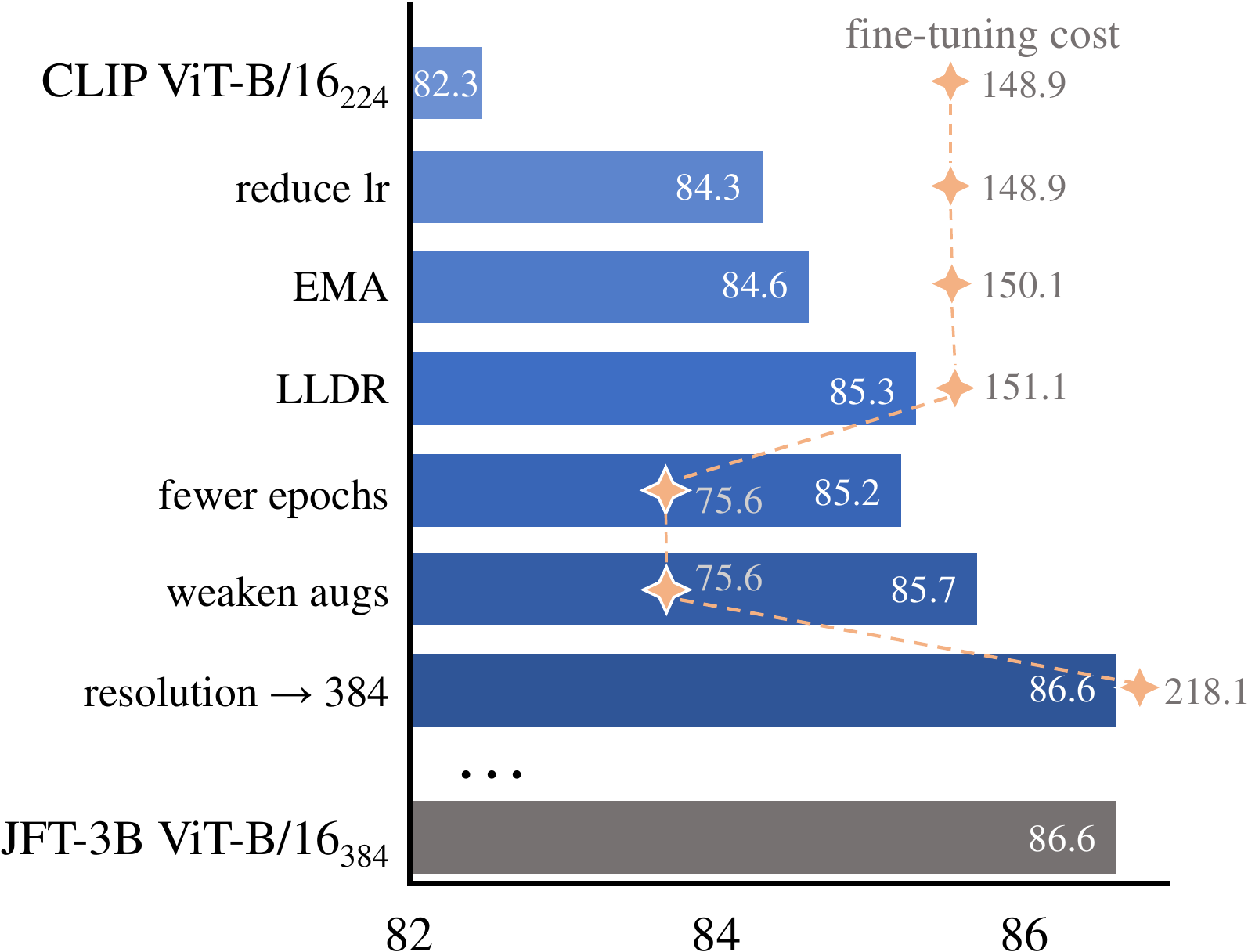}\\
	\vspace{-.7em}
	\caption{\textbf{Overview}. We show the components changed to improve the CLIP fine-tuning performance. With a proper fine-tuning strategy, the CLIP model  gets a comparable fine-tuning performance with the model supervisedly pre-trained on JFT.
		The ``fine-tuning cost'' denotes the GPU hours calculated with a single V100.
	}
	\label{fig:overview}
	\vspace{-1em}
\end{figure}

CLIP~\cite{radford2021learning} is undoubtedly of paramount importance in advancing the zero-shot performance for multiple computer vision tasks, especially for zero-shot image classification. 
By leveraging web-scale noisy image-text pairs crawled from the Internet and simply aligning the image features with text features via contrastive learning, CLIP demonstrates strong transferable ability and has been widely explored as cross-modal guidance in various multi-modal tasks~\cite{ramesh2021zero, ramesh2022hierarchical}.

Recently, masked image modeling (MIM) \cite{bao2021beit,xie2022simmim,wang2022bevt} has emerged as a successful visual pre-training paradigm and shows remarkable fine-tuning performance on downstream tasks. 
MIM essentially follows a mask-then-predict idea that masks a large ratio of the input image and forces the model to predict the masked part. 
The prediction target is a vital component and has attracted a lot of research efforts exploiting different targets such as discrete dVAE code~\cite{bao2021beit}, pixels~\cite{he2021masked, xie2022simmim}, perceptual codebook~\cite{dong2021peco}, HOG features~\cite{wei2021masked}, online features~\cite{dong2022boot, baevski2022data2vec,dong2022maskclip} and so on.
Until recently, it has been well acknowledged in many works~\cite{wei2022mvp,zhang2022cae,peng2022beit,fang2022eva} that utilizing CLIP features as prediction target of MIM leads to superior fine-tuning performance than other prediction targets.

Observing the success of distilling CLIP features to MIM models, we naturally ask a question: \emph{``If CLIP is a good teacher, why CLIP itself cannot be a strong fine-tuner?"} 
However, we notice that CLIP ImageNet fine-tuning results reported in previous works are significantly different from each other.
As a result, there is still not a clear answer to the question.
With this in mind, we investigate what is the real best performance that CLIP can achieve by directly fine-tuning itself on the ImageNet dataset (the CLIP image encoder to be more specific).

In this paper,
we justify that \emph{CLIP \textbf{itself}}, regardless of its impressive zero-shot capability, \emph{is super powerful under fine-tuning setting, achieving unprecedented performance and even outperforming all existing methods that use CLIP as the teacher}.
We provide concrete fine-tuning details in depth to ensure reproducibility.
Meanwhile, we also present 
a comprehensive analysis about each fine-tuning technique including the initial learning rate, the simultaneously updated EMA model, the layer-wise learning rate strategy, and fewer training epochs with weak input augmentation.
As shown in Fig.\ref{fig:overview}, 
starting from a commonly used ImageNet-1K fine-tuning setting,
we eventually end up with a proper fine-tuning strategy with several off-the-shelf fine-tuning techniques that significantly improve the CLIP fine-tuning performance.

The final results are surprisingly inspiring. 
With $224\times224$ input resolution, the CLIP ViT-Base model gets $85.7\%$ Top-1 accuracy on ImageNet-1K, even surpassing the state-of-the-art CLIP-targeted MIM method~\cite{peng2022beit} by $0.2\%$. With a larger input resolution of $384\times384$, the CLIP ViT-Base model achieves $86.6\%$ top-1 accuracy, comparable with the strong competitive with the same model size and same input resolution but is supervisedly trained over JFT-300M or even JFT-3B~\cite{zhai2022scaling}. 
We also provide fine-tuning results over a larger backbone ViT-Large, obtaining $88.0\%$ and $88.3\%$ top-1 accuracy with $224$ and $336$ input resolution respectively. 
We think these results are encouraging that, with the noisy web image-text data, CLIP could learn a very strong visual representation, which is comparable with the model learned from carefully annotated classification data at a similar scale, at least on the classification task.

In summary, this paper builds a strong and reproducible baseline for directly fine-tuning CLIP model itself.
Our paper demonstrates that the fine-tuning strategy is of crucial importance and justifies CLIP for ImageNet-1K fine-tuning. It will also motivate researchers in this field to rethink the latest proposed improvements upon CLIP.

\section{Experiments}
\subsection{Main Exp.}
We first report the baseline results. The backbone is initialized from the CLIP pretraining model and we add a new LayerNorm layer and a fully-connected layer as the classification head. We use the average pooling feature of the backbone output as the classification head input.  We finetune the model for 100 epochs, with the most commonly used augmentations and regularization, as shown in Table.\ref{tab:baseline}. We get a baseline result with $82.3\%$ top-1 ImageNet-1K classification accuracy. In the following, We gradually add or remove some of the finetune configuration components to further improve the performance.

\begin{table}[t]
	\tablestyle{6pt}{1.02}
	\footnotesize
	\begin{tabular}{l|l}
		config & value \\
		\shline
		optimizer & AdamW \\
		base learning rate & 1e-3 \\
		weight decay & 0.05 \\
		optimizer momentum & $\beta_1, \beta_2{=}0.9, 0.999$ \\
		batch size & 2048 \\
		learning rate schedule & cosine decay \\
		warmup epochs & 20 \\
		training epochs & 100\\
		augmentation & \texttt{RandAug} (m9, n2, mstd0.5) \cite{cubuk2020randaugment} \\
		label smoothing \cite{szegedy2016rethinking} & 0.1 \\
		mixup \cite{zhang2017mixup} & 0.8 \\
		cutmix \cite{yun2019cutmix} & 1.0 \\
		drop path \cite{Huang2016} & 0 \\
		random erase & 0.25, pixel \\
		random seed & 0 \\
		layer scale & No \\
		position encoding~(PE) & Learnable absolute PE \\
		\shline
		top-1 acc & 82.3 \\
	\end{tabular}
	\vspace{-.5em}
	\caption{\textbf{Baseline fine-tuning config and results for CLIP-Base.}}
	\label{tab:baseline}
	\vspace{-.5em}
\end{table}

\noindent\textbf{Adjust Learning Rate.} 
We first study the influence of different learning rates. As shown in the following Table \ref{table2}, we find that the fine-tuning need a quite small learning rate $2e^{-5}$ or $3e^{-5}$, about $50\times$ smaller than our default setting.

\begin{table}[h]\vspace{-3mm}
	\resizebox{1\linewidth}{!}{\setlength{\tabcolsep}{3mm}{
			\begin{tabular}{c|ccccccc}
				lr &  1e-5 &  2e-5 & 3e-5 & 5e-5 & 1e-4 & 5e-4 & 1e-3 \\
				\shline
				& 83.8 &  \textbf{84.3} & \textbf{84.3} & 83.8 & 83.1 & 82.3 & 82.3 \\
				
	\end{tabular}}}\vspace{-3mm}
	\caption{\textbf{Learning rate ablation.} A small learning rate improves performance greatly.}
	\vspace{-2mm}
	\label{table2}
\end{table}

\noindent\textbf{Exponential Moving Average (EMA).} 
Transferring a model pre-trained on a large dataset to a small new dataset always suffers from the overfitting problem, and the EMA is a commonly used method to relieve it. The EMA is realized by keeping a moving average of the weight of all the model parameters. We set the EMA momentum factor as $0.9998$ and report its influence with five different learning rates in Table \ref{table3}.
We find with the EMA, most of the settings get better results, especially when the learning rate is not proper, the EMA reduces its gap toward the best setting. 
\begin{table}[h]\vspace{-3mm}
	\resizebox{1\linewidth}{!}{\setlength{\tabcolsep}{.5mm}{
			\begin{tabular}{c|lllll}
				&  lr 1e-5 &  lr 2e-5 & lr 3e-5 & lr 5e-5 & lr 1e-4 \\
				\shline
				w/o EMA & 83.8 &  84.3 & 84.3 & 83.8 & 83.1 \\
				w EMA & 83.6~(-0.2) & 84.4~(+0.1) & \textbf{84.6}~(+0.3) & 84.4~(+0.6) & 84.0~(+0.9) \\
	\end{tabular}}}\vspace{-3mm}
	\caption{\textbf{Effectiveness of EMA.} EMA improves performance in most cases.}
	\vspace{-3mm}
	\label{table3}
\end{table}

\noindent\textbf{Layer-wise Learning Rate Decay (LLRD).}
The Layer-wise learning rate decay is a fine-tuning technique widely used in the Nature Language Processing~(NLP) area to fine-tune the BERT model. 
It has been used in BEiT \cite{bao2021beit} and shows significant performance.

The LLRD assigns different learning rates for each layer of the model backbone. It sets a large learning rate for the top layer and uses a multiplicative decay rate to decrease the learning rate layer-by-layer from top to bottom. 
With a large learning rate, the feature of the top layers changes more and could adapt to new tasks. On the contrary, the bottom layers have a small learning rate, so the strong feature learned from the pre-training is preserved.

We first pick the base learning rate (the learning rate of the last layer) from $\{5e^{-5}, 1e^{-4}\}$ and increase the LLDR from $0.6$ to $0.9$. As shown in Table \ref{table4}, we find a large base learning rate with a small LLDR performs better.
\begin{table}[h]\vspace{-3mm}
	\resizebox{1\linewidth}{!}{\setlength{\tabcolsep}{2.5mm}{
			\begin{tabular}{c|ccccccc}
				LLDR &  0.6  & 0.65  & 0.7  & 0.75 & 0.8 & 0.85  & 0.9 \\
				\shline
				lr 5e-5  & 84.6 & 84.7 & 84.8 & 84.9 & 84.7 & 84.6 & 84.4\\
				lr 1e-4  & 84.8 & 84.9 & \textbf{85.0} & \textbf{85.0} & 84.7 & 84.5 & 84.3 \\
	\end{tabular}}}\vspace{-3mm}
	\caption{\textbf{Layer-wise learning rate decay ablation.} A small decay factor with a large learning rate works better.}
	\label{table4}
\end{table}

Following the observation above, we pick LLDR from a small range $\{0.6, 0.65, 0.7\}$ and increase the learning rate from $1e^{-4}$ to $7e^{-4}$. As shown in Table \ref{table5}, we find with a proper LLDR,  all the results are better than our baseline result of $84.6\%$. With LLDR $0.6$ and learning rate $6e^{-4}$, we reach $85.3\%$ top-1 accuracy, $+0.7\%$ better than the EMA baseline.

\begin{table}[h]\vspace{-3mm}
	\resizebox{1\linewidth}{!}{\setlength{\tabcolsep}{2mm}{
			\begin{tabular}{c|ccccccc}
				lr &  1e-4  & 2e-4  &3e-4  &4e-4  &5e-4  &6e-4  &7e-4 \\
				\shline
				LLDR 0.60  & 84.8 & 84.8 & 85.0 & 85.1  & 85.2 & \textbf{85.3} & 85.1\\
				LLDR 0.65  & 84.9 & 84.9 & 85.0 & 85.0 & 85.0 & 84.9 & 84.9  \\
				LLDR 0.70  & 85.0 & 85.2 & 85.0 & ---  & --- & --- & --- \\
	\end{tabular}}}\vspace{-3mm}
	\caption{\textbf{Layer-wise learning rate decay ablation.} Further improve learning rate gets better results.}
	\label{table5}
\end{table}

\begin{figure}[t]\centering
	\vspace{-1em}
	\includegraphics[width=.95\linewidth]{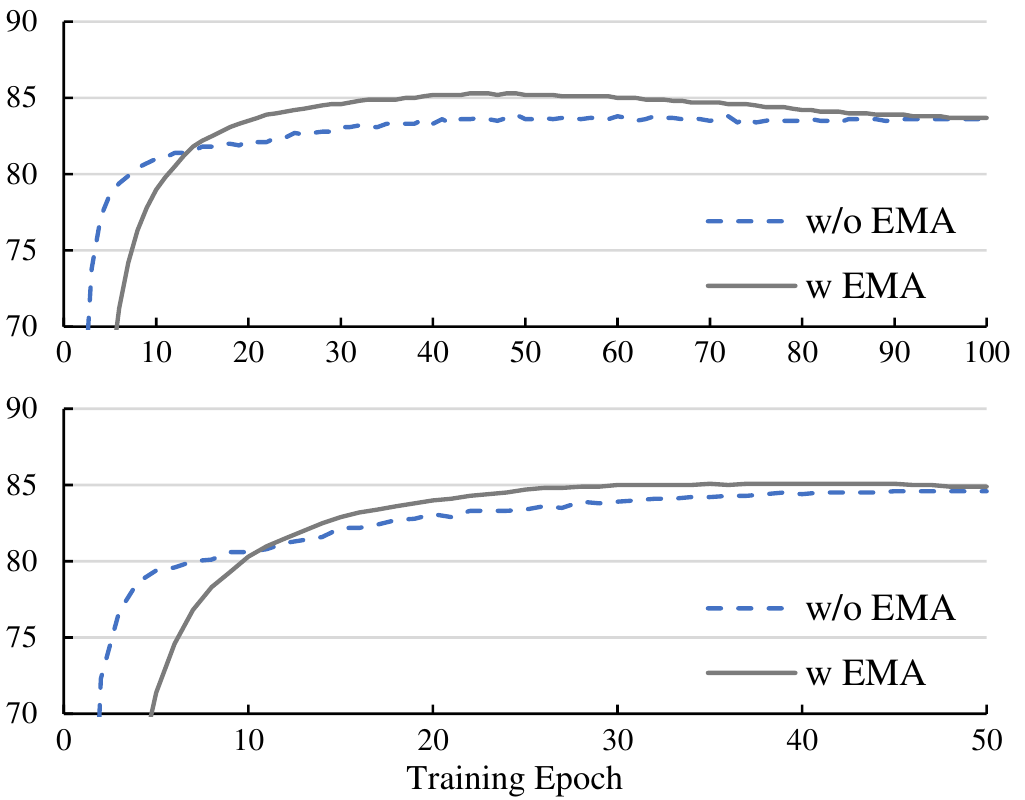}\\
	\vspace{-.7em}
	\caption{\textbf{Training Length}. Each figure shows the epoch-accuracy curve during the training.
		\textbf{Top}: 100 epoch fine-tuning setting, the model gets its best result with half of the training epochs and overfits the training set with the rest epochs.
		\textbf{Bottom}: 50 epoch fine-tuning setting, the model gets similar best accuracy and is under-fitting. 
	}
	\label{fig:epoch}
	\vspace{-1em}
\end{figure}

\noindent\textbf{Training Length.}
We show the epoch-accuracy curve in Fig.\ref{fig:epoch}, and we find that with 100 epoch fine-tuning, the model converges quite fast. It reaches the best performance before the 50$_{th}$ epoch and tends to overfit the training set with the rest epochs. 

When we reduce the training length to 50 epochs with 10 epochs warmup, and keep the rest setting unchanged. We find that the model gets a similar top-1 accuracy of $85.2\%$ and it seems under-fitting, as the online accuracy~(line w/o EMA in the figure) is still increasing. The shortened training length also reduces the fine-tuning cost to half, so we conduct the following experiments with the 50-epoch setting.

\noindent\textbf{Data Augmentation.}
In our default setting, we use two kinds of augmentations: content modification augmentations and content diversification augmentations. 
The former changes the image content with other images~(Mixup, CutMix), or removes part of the content~(Random Erase). 
The latter applies some affine transformation~(rotate, shear, translate, \etc in the Random Augmentation) or some statistics changes~(contrast, brightness, color, \etc in the Random Augmentation) to the image. 

We first study the influence of Mixup and CutMix in Table \ref{table6}. We find that under the 50 epoch setting, removing the Mixup/Cutmix gets the best results. We suspect that  as the feature learned by the CLIP model is good enough, it only needs some weak augmentations to transfer to a new dataset.

\begin{table}[h]
	\resizebox{1\linewidth}{!}{\setlength{\tabcolsep}{2mm}{
			\begin{tabular}{c|ccccc}
				MixUp/CutMix & 0.8/1.0  & 0.4/0.5  & 0.2/0.25 & 0.1/0.2 & 0.0/0.0 \\
				\shline
				&  85.3 & 85.4 & 85.5 & 85.6 & \textbf{85.7}\\
	\end{tabular}}}\vspace{-3mm}
	\caption{\textbf{MixUp\& CutMix Ablation.} Remove MixUP and CutMix gets better results.}
	\vspace{-3mm}
	\label{table6}
\end{table}

Then we follow the setting above and study the influence of Random Erase in Table \ref{table7}. We find that the influence of Random Erase is quite minor and the default setting works well.

\begin{table}[h]\vspace{-3mm}
	\resizebox{1\linewidth}{!}{\setlength{\tabcolsep}{4.5mm}{
			\begin{tabular}{c|cccc}
				Random Erase prob & 0.0  & 0.12  & 0.25 & 0.35 \\
				\shline
				&  85.6 & 85.6 & \textbf{85.7} & 85.5\\
	\end{tabular}}}\vspace{-3mm}
	\caption{\textbf{Random Erase Ablation.} The influence of random erase is minor.}
	\vspace{-3mm}
	\label{table7}
\end{table}

We further study the influence of Random Augmentation in Table \ref{table8}. It has two main hyper-parameter, $m$ for the augmentation magnitude (severity) and $n$ for the number of transformations selected per image.
When we fix $m=9$ and change $n$ from $1$ to $4$, we find $2$ or $3$ augmentations for each image work better.

\begin{table}[h]\vspace{-3mm}
	\resizebox{1\linewidth}{!}{\setlength{\tabcolsep}{6mm}{
			\begin{tabular}{c|cccc}
				n & 1  & 2  & 3 & 4 \\
				\shline
				m $= 9$ &   85.5 & \textbf{85.7} & 85.6 & 85.4\\
	\end{tabular}}}\vspace{-3mm}
	\caption{\textbf{RandAug augmentation magnitude $m$ Ablation.} Its influence is minor.}
	\vspace{-3mm}
	\label{table8}
\end{table}

Finally we fix $n=2$ and change the $m$ from $5$ to $10$~(10 is the max level). As shown in Table \ref{table9} We find the influence of the augmentation strength is quite small. 
\begin{table}[h]\vspace{-3mm}
	\resizebox{1\linewidth}{!}{\setlength{\tabcolsep}{4mm}{
			\begin{tabular}{c|cccccc}
				m & 5 & 6 & 7 & 8 & 9 & 10 \\
				\shline
				n $= 2$ &  85.6 & 85.6 & \textbf{85.7} & \textbf{85.7} & \textbf{85.7} & 85.6\\
	\end{tabular}}}\vspace{-3mm}
	\caption{\textbf{RandAug augmentation number $n$ Ablation.} 2 or 3 augmentations for each image works better.}
	\vspace{-3mm}
	\label{table9}
\end{table}

\begin{table}[t]
	\tablestyle{6pt}{1.02}
	\footnotesize
	\begin{tabular}{l|l}
		
		config & value \\
		\shline
		optimizer & AdamW \\
		base learning rate & 6e-4~(B), 4e-4~(L) \\
		Layer-wise lr decay & 0.6~(B), 0.65~(L) \\
		weight decay & 0.05 \\
		optimizer momentum & $\beta_1, \beta_2{=}0.9, 0.999$ \\
		batch size & 2048 \\
		learning rate schedule & cosine decay \\
		warmup epochs & 10~(B), 5~(L) \\
		training epochs & 50~(B), 30~(L)\\
		augmentation & \texttt{RandAug} (m9, n2, mstd0.5) \cite{cubuk2020randaugment} \\
		label smoothing \cite{szegedy2016rethinking} & 0.1 \\
		mixup \cite{zhang2017mixup} & 0 \\
		cutmix \cite{yun2019cutmix} & 0 \\
		drop path \cite{Huang2016} & 0 \\
		random erase & 0.25, pixel \\
		random seed & 0 \\
		EMA & 0.9998\\
		layer scale & No \\
		position encoding~(PE) & Learnable absolute PE \\
		\shline
		top-1 acc & 85.7~(B), 88.0~(L) \\
		
	\end{tabular}
	\vspace{-.5em}
	\caption{\textbf{Our fine-tuning config for CLIP-Base/Large.}}
	\label{tab:ours}
	\vspace{-.5em}
\end{table}

\noindent\textbf{Final Results.}
With the above attempts, we improve the CLIP-Base/16 fine-tuning accuracy from $82.3\%$ to $85.7\%$, with $+3.4\%$ improvement. 
We find all the successful attempts follow a similar idea: adapt the CLIP to new tasks while keeping its representation changes as slight as possible. For data augmentation, we remove the strong MixUp and CutMix which is totally different from the pre-training data, which may force the model to change more to adapt to it. For model parameter update, we shorten the training epochs, and use both layer-wise learning rate decay and EMA to slow down the change of the CLIP representation, especially its bottom layers representation.

In Table \ref{table11}, we compare our results with previous supervised pretraining methods\cite{steiner2021train,zhai2022scaling}, self-supervised MIM methods \cite{bao2021beit,he2021masked,chen2022context,dong2022bootstrapped,dong2021peco} and CLIP-based MIM methods~\cite{wei2022contrastive, zhang2022cae, peng2022beit, wei2022mvp} (i.e., use CLIP as the teacher). We find without any additional design nor training of a new model, the CLIP model with a proper fine-tuning setting outperforms these methods, surpassing the SOTA method by $0.2\%$.

We further apply the fine-tuning strategy to a larger resolution $384\times384$ and get $86.6\%$ top-1 accuracy.
Compared with the supervised training baseline with different data scales(results are copied from ~\cite{steiner2021train} and appendix of ~\cite{zhai2022scaling}, selecting best results for each setting), we find that the CLIP with 400M web image-text pairs shows similar performance with the carefully annotated JFT dataset with similar or even larger data scale. 

We also apply the fine-tuning strategy to the CLIP-Large/14 model. With the configurations shown in Table.\ref{tab:ours}, we get a strong result, $88.0\%$ top-1 accuracy with $224\times224$ resolution input. For the ViT-Large model, the CLIP use ViT-L/14 with $224\times224$ resolution input while the previous works report results of ViT-L/16 with a larger input $384\times384$. Here we show their FLOPs for clear comparison. We find that with half of the FLOPs, the CLIP with ViT-L/14$_{224}$ shows comparable results with the JFT-300M pre-trained model. When we further increase the input resolution to $336\times336$~(190.6G FLOPs, similar to the ViT-L/16$_{384}$ with 190.7G FLOPs), we get $88.3\%$ top-1 accuracy, better than the model trained on JFT-300M result and slightly worse than that on much larger data JFT-3B.

We think these results are encouraging that only with the noisy image-text data collected from the Internet, the CLIP could learn a very strong visual representation, which is comparable with the model learned from carefully annotated classification data at a similar scale, at least on the classification task. This shows the strong capability and potential of contrastive image-language learning.

\begin{table}\vspace{-.2em}
	\setlength{\tabcolsep}{0.5mm}{
			\begin{tabular}{l|cc|ccc}
				\toprule
				& B/16$_{224}$ & B/16$_{384}$  & L/16$_{384}$ & L/14$_{224}$ & L/14$_{336}$\\
				\midrule
				FLOPs & 17.5G & 55.4G & 190.7G & 80.7G & 190.6G \\
				\midrule
				\multicolumn{6}{l}{\textit{Supervised Baseline}} \\
				IN-21K~\cite{steiner2021train}   &  84.0  & 86.2 & 87.1 &--- &---\\
				JFT-300M~\cite{zhai2022scaling} &  ---   & 86.7 & 88.0 & --- &--- \\
				JFT-3B~\cite{zhai2022scaling} &  ---   & 86.6   & 88.5 & --- &--- \\
				\midrule
				\multicolumn{6}{l}{\textit{Self-supervised MIM}} \\
				BEiT\cite{bao2021beit}    & 83.2 & ---  &--- & ---  & ---  \\
				MAE\cite{he2021masked}     & 83.6 & ---  &--- & ---  & --- \\
				CAE\cite{chen2022context}   & 83.8 & ---  &--- & ---  & ---  \\
				BootMAE\cite{dong2022bootstrapped} & 84.2 & ---  &--- & ---  & --- \\
				PeCo\cite{dong2021peco}   & 84.5 & ---  &--- & ---  & ---  \\
				\midrule
				\multicolumn{6}{l}{\textit{MIM with CLIP as prediction target}} \\
				MVP\cite{wei2022mvp}     & 84.4 & ---  &--- & ---  & ---  \\
				FD-CLIP\cite{wei2022contrastive} & 84.9 & ---  &--- & 87.7 & --- \\
				CAE-v2\cite{zhang2022cae}  & 85.3 & ---  &--- & ---  & ---  \\
				BEiT-2\cite{peng2022beit}  & 85.5 & ---  &--- & ---  & ---  \\
				\midrule
				\multicolumn{6}{l}{\textit{Fine-tuning CLIP directly}} \\
				Ours &  85.7 & 86.6 & ---  & 88.0 & 88.3 \\
				\bottomrule
	\end{tabular}
	}\vspace{-3mm}
	\caption{\textbf{Comparisons with previous results on ImageNet1K.} We compare with supervised baseline trained with different dataset scales, and MIM-based methods with different prediction targets, especially the methods using CLIP as the prediction target. We find CLIP itself is a strong fine-tuning model, outperforming existing methods.}
	\label{table11}
\end{table}

\subsection{More Ablations.}

\begin{figure}[t]\centering
	\vspace{-1em}
	\includegraphics[width=1\linewidth]{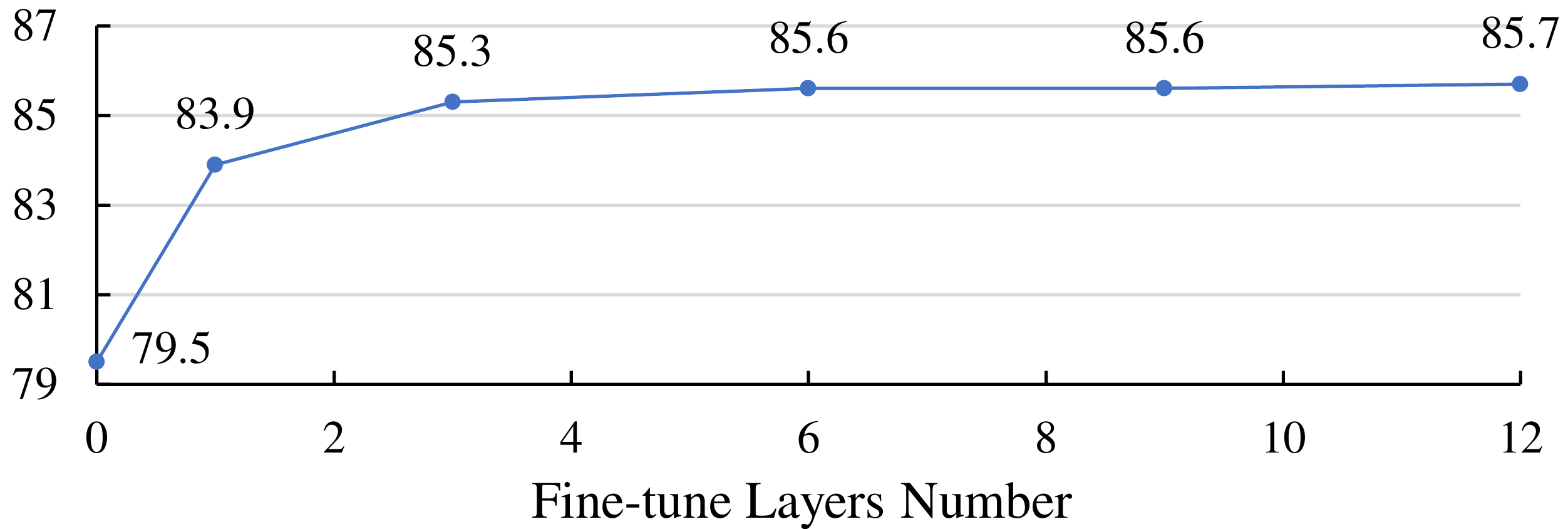}\\
	\vspace{-.7em}
	\caption{\textbf{Partial fine-tuning} results of CLIP-Base/16. The 0 layer tuning is linear probing and 12 is the full fine-tuning. The feature learned by CLIP is quite strong that freezing half of the layers gets $85.6\%$ top-1 accuracy, close to the full fine-tuning result.
	}
	\label{fig:layers}
	\vspace{-1em}
\end{figure}
\noindent\textbf{Partial fine-tuning.}
The layer-wise learning rate decay is used with an idea that the early layers learn quite good features from the large-scale pre-training so it only needs a small learning rate and changes slightly. Here we use the partial fine-tuning design to study the feature learned from different layers. In practice, we fine-tune part of the model layers and freeze the rest layers. 
As shown in Fig.\ref{fig:layers}, with only one layer fine-tuning, the model gets $83.9\%$ top-1 accuracy, this shows the feature learned by CLIP is strong. With half of the layers frozen, the model gets $85.6\%$ top-1 accuracy, close to the fully fine-tuning results. This also proves the necessity of the layer-wise learning rate decay that the bottom layers only need small changes.

\noindent\textbf{Architecture Modification.} 
CLIP uses the vanilla Vision Transformer architecture with absolute position encoding~(APE), and there are several advanced techniques to further improve the model performance, such as the relative position encoding~(RPE) and layer scale~(LS). 
The RPE provides more inductive bias with a relative position table, and LS adds a learnable scale factor for each block, improving the model capability.
Here we further add them to the fine-tuning model and study their influence in Table \ref{table12}. In practice, we initialize the RPE table with $0$ and the LS factor with $1$. However, their influence is quite minor. This may be because the model already learns enough inductive bias through large-scale pre-training. This is also consistent with the observation in ~\cite{raghu2021vision} that a vanilla ViT trained with large-scale data could learn a similar pattern to a hybrid ViT that uses conv-stem to attend local manually.

\begin{table}[h]\vspace{-3mm}
	\resizebox{1\linewidth}{!}{\setlength{\tabcolsep}{6mm}{
			\begin{tabular}{c|ccc}
				architecture &  vanilla  & + RPE & +RPE +LS \\
				\shline
				& \baseline{85.66} & 85.69 & 85.67 \\
				
	\end{tabular}}}\vspace{-3mm}
	\caption{\textbf{Model Architecture Ablation.} Additional components added in the fine-tuning are not helpful. Setting with \gc{gray} background is our default setting.}
	\vspace{-3mm}
	\label{table12}
\end{table}

\noindent\textbf{Data Augmentations.} 
In our main experiment, we study the influence of commonly used data augmentations. Here we further study two other augmentations in Table \ref{table13}: 3Aug~\cite{touvron2022deit} and the AugMix\cite{hendrycks2019augmix}. The 3Aug also proposes using single resize crop~(SRC), instead of the default random resize crop~(RRC).
We find that the 3Aug+RRC performs similarly to the RandAug, while the 3Aug+SRC and AugMix perform worse. 
\begin{table}[h]\vspace{-.2em}
	\resizebox{1\linewidth}{!}{\setlength{\tabcolsep}{2mm}{
			\begin{tabular}{c|cccc}
				Augs & RandAug+RRC & 3Aug+RRC & 3Aug+SRC & AugMix+RRC \\
				\shline
				& \baseline{85.7} & 85.6 & 84.9 & 85.1 \\
	\end{tabular}}}\vspace{-3mm}
	\caption{\textbf{Augmentation Method Ablation.} The default RandAug with random resize crop~(RRC) works well.}
	\vspace{-3mm}
	\label{table13}
\end{table}

In the random resize crop, an important factor is the crop ratio, it controls the cropped region size of the input image. Its upper bound is 1 and the default lower bound is 0.08. A larger lower bound leads to a weaker augmentation that the cropped results contain more area of the input. Here we study the influence of the lower bound. We find the influence is minor and a smaller crop ratio works better.

\begin{table}[h]\vspace{-.2em}
	\resizebox{1\linewidth}{!}{\setlength{\tabcolsep}{4.5mm}{
			\begin{tabular}{c|ccccc}
				crop ratio & 0.08 & 0.2 & 0.3 & 0.4 & 0.5 \\
				\shline
				& \baseline{85.7} & 85.7 & 85.6 & 85.6 & 85.5 \\
	\end{tabular}}}\vspace{-3mm}
	\caption{\textbf{Random Resize Crop Ratio Ablation.} The influence is minor and a smaller crop ratio works better.}
	\vspace{-3mm}
\end{table}

\noindent\textbf{Model regularization.} 
Here we study the influence of different regularization methods.

\textbf{Drop path rate}~(DPR) is a commonly used regularization to ease the over-fitting problem. For vision transformer architecture, it randomly drops the MHSA or FFN results in each block and only output the shortcut result. In most cases, we set the transfer learning DPR slightly larger than it was used in the pretraining to get better results.
The CLIP is pre-trained with DPR$=0$ and here we study the fine-tuning performance with a larger DPR. As shown in the following Table \ref{table15}, the fine-tuning performance decrease with using DPR. This may be because the DPR is not used in the CLIP pre-training so the model is not aware of such regularization.
\begin{table}[h]\vspace{-.2em}
	\resizebox{1\linewidth}{!}{\setlength{\tabcolsep}{9mm}{
			\begin{tabular}{c|ccc}
				DPR &  0  & 0.05 & 0.1 \\
				\shline
				& \baseline{85.7} & 85.6 & 85.5 \\
				
	\end{tabular}}}\vspace{-3mm}
	\caption{\textbf{Drop Path Rate Ablation.} The DPR is not used in the pre-training, increasing it in the fine-tuning leads to worse results.}
	\vspace{-3mm}
	\label{table15}
\end{table}

\textbf{Label smoothing} is a regularization used in the cross-entropy loss to solve the overconfident problem of the classification model. As shown in Table \ref{table16}, we find that removing it leads to overfitting and worse performance.
\begin{table}[h]\vspace{-.2em}
	\resizebox{1\linewidth}{!}{\setlength{\tabcolsep}{3mm}{
			\begin{tabular}{c|ccccc}
				Label smoothing &  0  & 0.05 & 0.1 & 0.15 & 0.2 \\
				\shline
				& 85.3 & 85.6 & \baseline{85.7} & 85.6 & 85.6  \\
	\end{tabular}}}\vspace{-3mm}
	\caption{\textbf{Label Smoothing Ablation.} Removing label smoothing leads to worse results.}
	\vspace{-3mm}
	\label{table16}
\end{table}

\textbf{Weight Decay} is an additional loss calculated in the optimizer to prevent overfitting. In Table \ref{table17}, we find that its influence is quite minor.
\begin{table}[h]\vspace{-.2em}
	\resizebox{1\linewidth}{!}{\setlength{\tabcolsep}{4mm}{
			\begin{tabular}{c|ccccc}
				weight decay & 1e-8 & 0.01 & 0.05  & 0.1 & 0.2 \\
				\shline
				& 85.6 & 85.6 & \baseline{85.7} & 85.6 & 85.6  \\
	\end{tabular}}}\vspace{-3mm}
	\caption{\textbf{Weight Decay Ablation.} Its influence is negligible.}
	\vspace{-3mm}
	\label{table17}
\end{table}

\section{Conclusion}
In this paper, we study how far the performance can be achieved by directly  fine-tuning the CLIP on the ImageNet-1K dataset.
We find that CLIP itself is a super strong finetuner, even outperforming existing methods that use CLIP as the teacher. We hope that our study could be a new baseline for the following works, raise attention to the strong recognition capability of the CLIP model, and rethink recent improvements based on CLIP. In the future, we will transfer our strategy to other multimodal vision foundation models Florence\cite{yuan2021florence} and OmniVL\cite{wang2022omnivl}.

{\small
	\bibliographystyle{ieee_fullname}
	\bibliography{egbib}
}

\end{document}